\documentclass[10pt,twocolumn,letterpaper]{article}

\pdfoutput=1

\usepackage{cvpr}
\usepackage{times}
\usepackage{epsfig}
\usepackage{graphicx}
\usepackage{amsmath}
\usepackage{amssymb}

\input{macros}

\usepackage[breaklinks=true,bookmarks=false]{hyperref}

\cvprfinalcopy 

\begin{document}

\title{Learning Complete 3D Morphable Face Models from Images and Videos}

\author{ Mallikarjun B R \quad Ayush Tewari \quad Hans-Peter Seidel \quad Mohamed Elgharib \quad Christian Theobalt \\
 Max Planck Institute for Informatics, Saarland Informatics Campus }

\maketitle

\begin{abstract}

Most 3D face reconstruction methods rely on 3D morphable models, which disentangle the space of facial deformations into identity geometry, expressions and skin reflectance. 
These models are typically learned from a limited number of 3D scans and thus do not generalize well across different identities and expressions. 
We present the first approach to learn complete 3D models of face identity geometry, albedo and expression just from images and videos. 
The virtually endless collection of such data, in combination with our self-supervised learning-based approach allows for learning face models that generalize beyond the span of existing approaches. 
Our network design and loss functions ensure a disentangled parameterization of not only identity and albedo, but also, 
for the first time, an expression basis.  
Our method also allows for in-the-wild monocular reconstruction at test time. 
We show that our learned models better generalize and lead to higher quality image-based reconstructions than existing approaches. 
\end{abstract}

\section{Introduction}
\label{sec:intro}

Monocular 3D face reconstruction is defined as recovering the dense 3D facial geometry and skin reflectance of a face from a monocular image. 
It has applications in several domains such as VR/AR, entertainment, medicine, and human computer interaction. 
We are concerned with in-the-wild images which can include faces of many different identities with varied expressions and poses, in unconstrained environments with widely different illumniation. 
This problem has been well-studied, where a lot of success can be owed to the emergence of \emph{3D Morphable Models}~\cite{Blanz1999}. 
These morphable models define the space of deformations for faces as separate disentangled models such as facial identity, expression and reflectance. 
These models are widely used in the literature to limit the search space for reconstruction~\cite{Zollhoefer2018FaceSTAR,egger20193d}.
However, these morphable models are often learned from a limited number of 3D scans, which constrains their generalizability to subjects and expressions outside the space of the scans.  

\begin{figure}
\centering
	\includegraphics[width=1\linewidth]{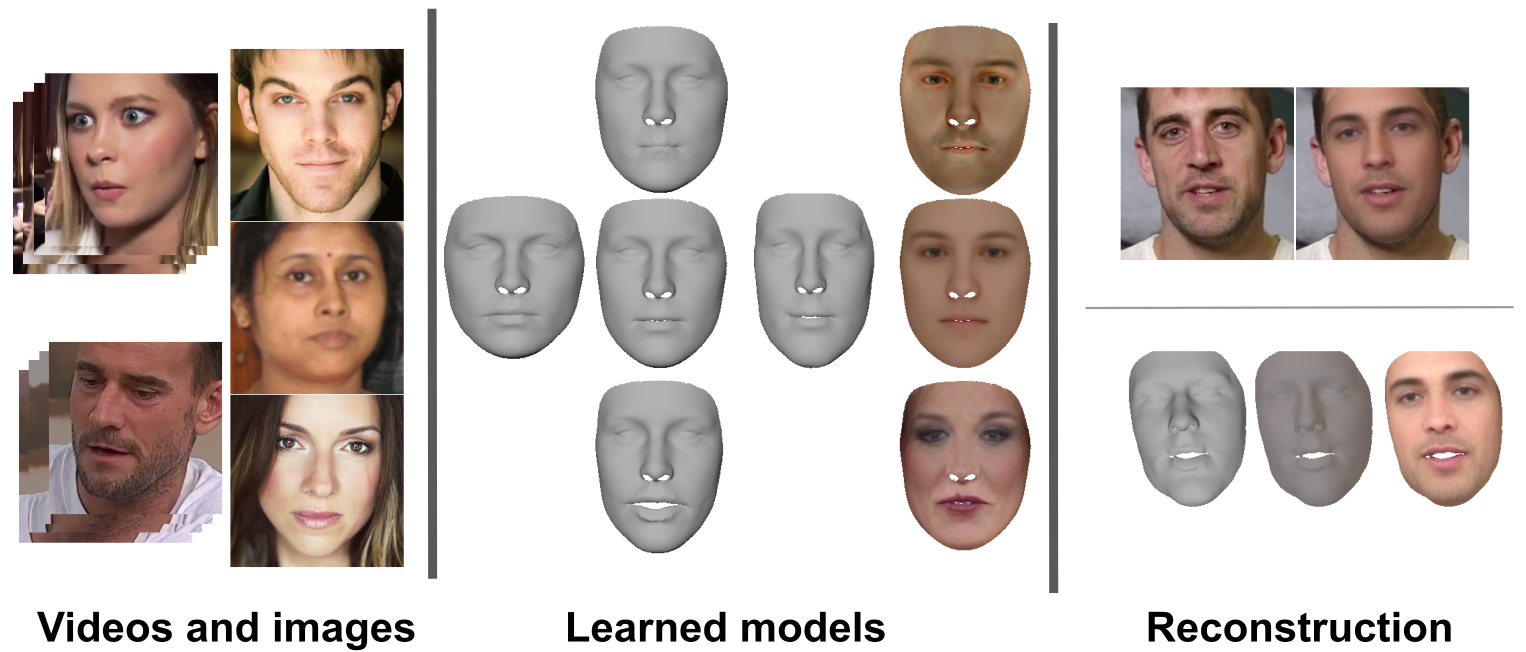}
	\caption{We present the first approach for learning complete 3D morphable face models from in-the-wild images. Our approach learns models for identity geometry, expression and albedo, completely in a self-supervised manner. It achieves good disentanglement of the various facial components and  produces high quality photorelism in the final overlay.}
	\label{fig:Teaser}
\end{figure}
Recent efforts examined learning face models with better generalizability from internet images or videos \cite{Tewari19FML,tewari2018self,Tran19,Tran2018b,Tran18}. %
However, learning from in-the-wild data is highly challenging, requiring solutions for handling the strong inherent ambiguities and for ensuring disentanglement between different components of the reconstruction. 
Some approaches deal with a slightly easier problem of refining an initial morphable model pretrained on 3D data on in-the-wild imagery~\cite{tewari2018self,Tran2018b,Tran2017,Tran18,Tran19}. 
Our objective is to learn face models without using any pretrained models to start with. 
The closest approach to ours is Tewari~\etal~\cite{Tewari19FML}, which learns only the models of facial identity geometry and reflectance from in-the-wild videos. 
However, they still use a pretrained expression  model to help disentangle the identity and expression variations in geometry. 
We present the first approach that learns the the complete face model of identity geometry, albedo and expression from just from in-the-wild videos. 
We start just from a template face mesh without using any priors about deformations of the face, other than smoothness. 
This also makes ours the first approach to learn face expression models from 2D data.

We achieve this through several technical contributions. 
We design a neural network architecture which, in combination with specially tailored self-supervised loss functions, enables (1) learning of face identity, expression and skin reflectance models, as well as (2) joint 3D reconstruction of faces from monocular images at state-of-the-art accuracy. 
We use a siamese network architecture which can process multiple frames of video during training, and enables consistent identity reconstructions and expression basis reconstruction.%
We use a differentiable renderer to render synthetic images of the network's reconstructions.
To compare reconstructions to the input, we use a new combination of appearance-based and face segmentation losses that permit learning of overall face geometry and appearance, as well as a high-quality expression basis of detailed mouth and lip motion.
Our novel lip segmentation consistency loss aligns the lip region in 3D with 2D segmentations. 
Our loss is robust to noisy outliers, leading to qualitatively better lip segmentations than the ground truth used. 
We also introduce a disentanglement loss which ensures that the expression component of a reconstructed mesh is small when the input image contains a neutral face. 
We show that the combination of these innovations is crucial to learn a full face model with proper component disentanglement from in-the-wild imagery, and out performs the state-of-the-art image-based face reconstruction. 

In summary we make the following contributions: 1) the first approach for learning all components - identity, albedo and expression bases - of a morphable face model, trained on in-the-wild 2D data, 2) the first approach to learn 3D expression models of faces in a self-supervised manner, 3) a lip segmentation consistency loss to enforce accurate mouth modeling and reconstruction, 4) enforcing disentanglement of identity and expression geometry by utilizing a dataset of neutral images.

\section{Related Work}
\label{sec:related}

\subsection{Face Modeling}
Faces are typically modeled as a combination of several components: reflectance, identity geometry and expression. 3D parametric identity~\cite{Blanz1999,Blanz2003} and blendshapes~\cite{Pighin98,Lewis14,Tena11} are  used to represent identity (geometry and reflectance) and facial expressions. This generalizes active appearance models~\cite{CooteET2001} from 2D to 3D space. To model the variations among different people, a parametric PCA space can be learned from a dataset of laser scans~\cite{Blanz1999,Blanz2003}. This represents deformations in a low-dimensional subspace. The resulting 3D morphable face model (3DMM) is the most widely used face model in literature~\cite{Zollhoefer2018FaceSTAR}. Multi-linear face models extend this base concept; they often use tensor-based representations to better model the correlations between shape identity and expression~\cite{DaleSJVMP2011,Bolkart16,Abrevaya18}.

Physics-based face models~\cite{Ichim17,Sifakis05} have been proposed, however their complexity makes their use in real-time rendering or efficient reconstruction difficult. Animation artists can also manually create face rigs, with custom-designed control parameters. They often use blendshapes, linear combinations of designed base expressions, to control face expressions~\cite{Lewis14}.  
Recently, large collections of 3D and 4D (3D over time) scans have been used to learn face models. In~\cite{BoothRPDZ18} thousands of 4D scans are used to learn a parametric face model. Li~\etal~\cite{Li17} used 33,000 3D scans to learn the FLAME face model. The model combines a linear shape space with articulated motions and semantic blendshapes. 

\subsection{Face Reconstruction}

Image-based reconstruction methods~\cite{Zollhoefer2018FaceSTAR} estimate face reflectance, geometry and/or expressions. 3DMMs~\cite{Blanz1999,Blanz2003} is often used as priors for this task. Methods differ in the type of input they use, such as monocular~\cite{Romdhani05}, multi-frame~\cite{Tewari19FML} or unstructured photo collection input~\cite{Roth:2016}. Current methods can be classified into 1) optimization-based and 2) learning-based. Optimization-based techniques rely on a personalized model~\cite{Cao16,Fyffe14,GVWT13,Wu16b,Ichim15} or a parametric prior~\cite{Agudo14,Bouaziz13,Li13,Shi14} to estimate 3D geometry, often combined with texture and/or illumination, from a 2D video. 

Learning-based approaches regress 3D face geometry from a single image by learning an image-to-parameter or image-to-geometry mapping~\cite{olszewski2016high,Richardson_2017_CVPR,tewari17MoFA,tewari2018self,sela2017unrestricted,Tran_2017_CVPR,KimZTTRT17}. These methods require ground truth face geometry~\cite{Tran_2017_CVPR,Laine:2017}, synthetic data generated from a morphable model~\cite{RichaSK2016,Richardson_2017_CVPR,sela2017unrestricted,KimZTTRT17}, or a mixture of both~\cite{McDonagh2016,Klaudiny:2017,Tran2017}. Tewari et al.~\cite{tewari17MoFA} propose a differentiable rendering-based loss which allows  unsupervised training from 2D images. Genova~\etal~\cite{Genova2018} learn to regress an image into 3D morphable model coordinates using unlabelled data. They impose identity similarity between the input and the output,
in addition to a loop-back loss and a multi-view identity loss. Deng~\etal~\cite{deng2019} combined image-consistency and perceptual loss which leads to improved results.  
A new multi-image shape confidence learning scheme is also proposed which outperforms naive aggregation and other heuristics. 
\emph{RingNet}~\cite{RingNet19} estimates the parameters of the FLAME model~\cite{Li17}. It utilizes multiple images during training, and enforces the shape to be the same for pictures of the same identity, and different for different people. 
While  these techniques are fast and produce good results, reconstructing shape and appearance variations outside the pre-dedfined 3DMM space is difficult. 

\begin{figure*}
    \centering
    \includegraphics[width=0.9\linewidth]{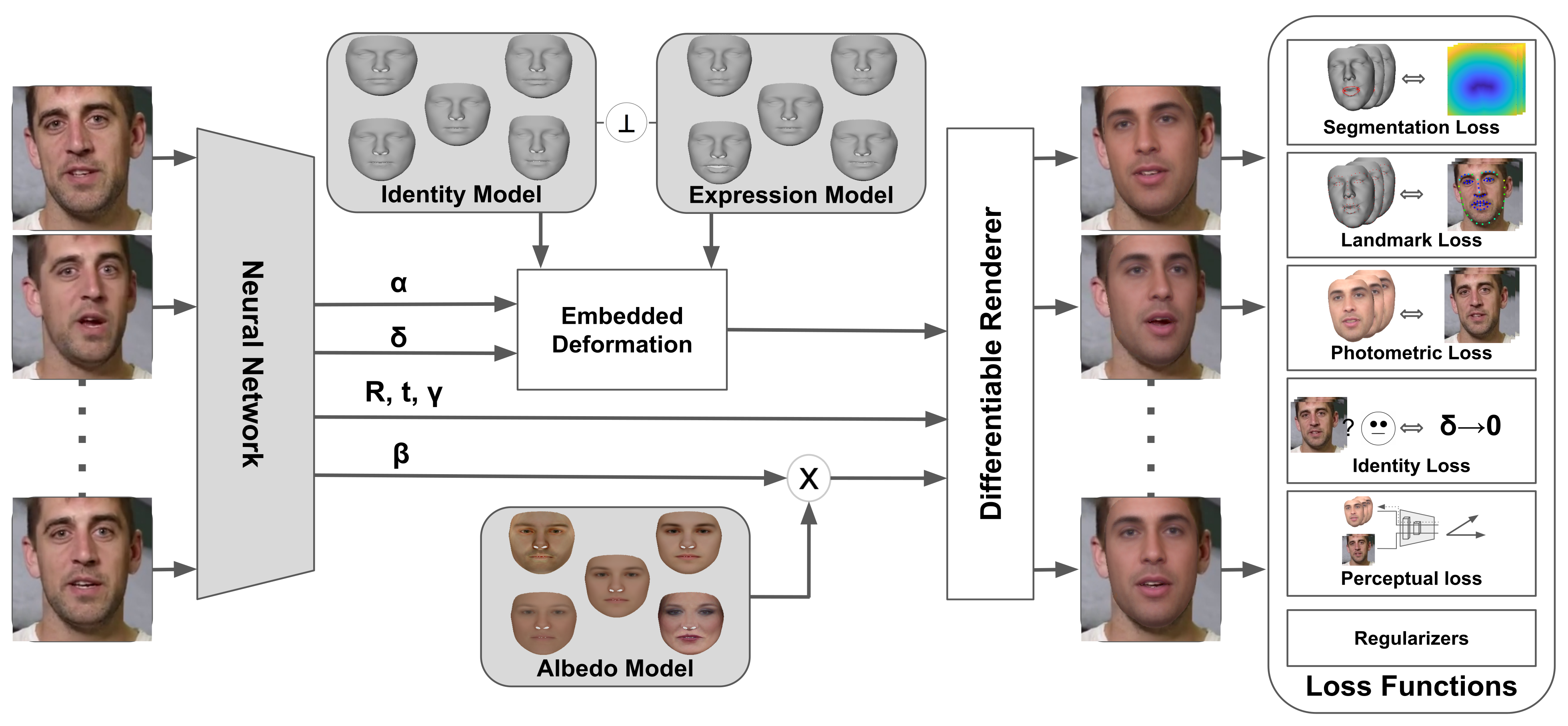}
    \vspace{-0.1cm}
    \caption{Our approach jointly learns identity, expression and albedo models along with the input-dependent parameters for these models. The network is trained in a siamese manner using differentiable renderer to compute self-supervised loss.}
    \label{fig:pipeline}
    \vspace{-0.2cm}
\end{figure*}
\subsection{Joint Modeling and Reconstruction}
Recent learning-based methods for monocular face reconstruction~\cite{tewari2018self,Tran18,Tran2018b,Booth_2017_CVPR,Sengupta18,Tewari19FML} allow for capturing variations outside of the 3DMM space by  training from in-the-wild data. Tran~\etal~\cite{Tran18} employ two separate convolutional decoders to learn a non-linear model that disentangles shape from appearance. Similarly, Sengupta~\etal~\cite{Sengupta18} propose residual blocks to produce a complete separation of surface normal and albedo features. 
Tewari~\etal~\cite{tewari2018self} learn a corrective space of albedo and geometry which generalizes beyond 3DMMs trained on 3D data.
Lin~\etal~\cite{lin2020towards}  refine the initial texture generated by the 3DMM albedo model. Lee ~\etal~\cite{Lee_2020_CVPR} learn a non-linear identity model, with a fixed existing expression model.
Learning morphable models from scratch in a relatively less studied problem. 
Tewari~\etal~\cite{Tewari19FML}  learn the identity (shape and reflecance) model from community videos in a self-supervised manner. 
The learning starts from a neutral reflectance and coarse deformation graph, which are refined during training. 
It however relies on a learned expression model.
Our method is the first to learn all dimensions--reflectance, identity geometry, and expression from in-the-wild data.

\section{Method}
\label{sec:main}

We present the first method to learn a deformable face model that jointly learns all three of the following dimensions - identity geometry, expression and reflectance - from unlabelled community videos, without using a pre-defined 3DMM to start with. 
The starting point for our deformation models is a mesh which defines the topology of reconstructions, as well as the initial geometry and reflectance values for our networks.
We design a multi-frame siamese network which processes the videos at training time. 
The training is self-supervised, without any 3D supervision. 
We use a differentiable renderer to define our loss functions in the image space.
Our network design, in addition to loss functions enable disentangled learning of the face model subspaces. 
Our network also jointly learns to predict parameters of the models, thus enabling 3D reconstruction at test time, even from monocular images. 
\subsection{Model Representation}

We learn a linear face models, similar to many existing face models~\cite{Blanz1999,tewari2018self,Tewari19FML}, comprising linear models of identity geometry, expression geometry and albedo. 
(Stacked) Mesh vertex positions and reflectances are represented as $V$ and $R$, $|V|=|R|=3N$, where $N$ is the number of vertices.
We use the mesh topology of Tewari~\etal~\cite{tewari2018self} with $N=60,000$ vertices.\\
\paragraph{Geometry Models}
3D face deformations due to identity and expression can be represented using linear geometry models. 
\begin{equation}
\label{eq:model_geometry}
    V(\matM_{id}, \matM_{exp}, \alpha, \delta) = \bar{V} + \matM_{id} \alpha + \matM_{exp} \delta
\end{equation}
Here, $\matM_{id} \in \mathbb{R}^{3N \times m_i}$ and $\matM_{exp} \in \mathbb{R}^{3N \times m_e}$ are the learnable linear identity and expression models.  We use the mean face from~\cite{Blanz2004} as $\bar{V}$. 
$\alpha \in \mathbb{R}^{m_i}$ and $\delta \in \mathbb{R}^{m_e}$ are the identity and expression parameters for the corresponding models. 

We use a low-dimensional embedded deformation graph to represent the linear models $\matM_{id}$ and $\matM_{exp}$
\begin{gather}
    \matM_{id} = \matU \matM_{gid}, 
    \matM_{exp} = \matU \matM_{gexp} 
\end{gather}
Here, $\matM_{gid} \in \mathbb{R}^{3G \times m_i}$ and $\matM_{gexp} \in \mathbb{R}^{3G \times m_e}$ are linear models defined on a lower dimensional graph with $G = 521$ nodes.
The fixed upsampling matrix $\matU \in \mathbb{R}^{3N \times 3G}$ couples the deformation graph to the full face mesh and is precomputed before training. 
Learning the shape models in the graph-space reduces the number of learnable parameters in the model, and makes it easier to formulate smoothness constraints over the reconstructions.\\
\paragraph{Reflectance Model}
We employ a linear model of diffuse face reflectance. 
\begin{equation}
\label{eq:model_ref}
    R (\matM_R, \beta) = \bar{R} + \matM_R \beta
\end{equation}
Here, $\matM_R \in \mathbb{R}^{3N \times m_r}$ is the learnable reflectance model, and $\beta \in \mathbb{R}^{m_r}$ are the estimated parameters.
We use the mean face reflectance from~\cite{Blanz2004} as $\bar{R}$.
Unlike geometry, we learn a per-vertex reflectance model on the full mesh resolution. 
This allows us to preserve photorealistic details of the face in the reconstructions.

\subsection{Image Formation}
Given a face mesh with positions $V$ and reflectance values $R$, we additionally need the extrinsic camera parameters in order to render synthetic images. 
Rigid face pose is represented as $\phi(v) = Rot(v) + t$, where $t$ includes $3$ translation parameters, and rotation $Rot\in SO(3)$ is represented in $3$ Euler angles. 
We use a perspective camera model, with projection function $\pi:\mathbb{R}^3\xrightarrow{}\mathbb{R}^2$. 
For any point $v\in\mathbb{R}^3$, the corresponding projection $p(v)\in\mathbb{R}^2$ is defined as $p(v)=\pi(\phi(v))$.

To define the color, we need to model the scene illuimination. 
We assume a lambertian surface, and use spherical harmonics (SH) coefficients $\gamma$ to represent the illumination~\cite{Ramamoorthi2001}. 
The color $c$ of a point with reflectance $r$ and position $v$ can be computed as
\begin{equation}
    c = r \cdot \sum_{b=1}^{B^2} \gamma_b \cdot \textbf{H}_b(n)
\end{equation}
$\textbf{H}_b: \mathbb{R}^3 \xrightarrow{} \mathbb{R}$ are the SH basis functions, $\gamma\in\mathbb{R}^{B^2}$ are the SH coefficients, $n$ are the normals at point $v$ and $B=3$ .

\paragraph{Differentiable Rendering}
We implement a differentiable rasterizer to render 2D images from 3D face meshes. 
For each pixel, we first compute the 3D face points which project into the pixel. 
We use a z-buffering algorithm to select the visible triangles. 
Pixel color is computed by linearly interpolating between vertex colors using barycentric coordinates. 
We implement the renderer in a data-parallel fashion as a custom TensorFlow layer. 

This implementation also allows for gradients to back propagate through the rendering step. 
The gradients computed at any pixel location can be distributed across the vertices of the relevant triangle according to the barycentric coordinates. 
While such an implementation cannot differentiate through the visibility check, it is appropriate in practice. 

\label{sec:image_formation}
\subsection{Network Architecture}
Our network consists of siamese towers which take as input different frames of a video $F_i$ $\forall i\in[0,{N_f}]$, where ${N_f}$ is the number of frames. 
Each such set of ${N_f}$ frames of one person identity is called a multi-frame image.
The output of the siamese towers are the face parameters which are independent per-frame, i.e. expressions ($\delta_i$), illumination ($\gamma_i$) and rigid pose ($\phi_i$) $\forall i\in[0,{N_f}]$.
We formulate multi-frame constraints for the identity component of the model. 
By design, the network only produces one output per multi-frame input for the identity shape ($\alpha$) and reflectance ($\beta$).
This is done through a multi-frame pooling of features from the siamese towers, followed by a small network. 
Thus, the network produces per-frame parameters, $\vecp_i = (\alpha, \beta, \delta_i, \gamma_i, \phi_i)$ 

In addition to the face parameters, we also learn the face models for expression ($\matM_{exp}$), identity shape ($\matM_{id}$) and albedo ($\matM_R$).
These models are implemented as weights of the learnable network. 
More specifically, the position and reflectance of the face mesh, represented as $V_i(\matM_{id},\matM_{exp},\alpha_i,\delta_i$) and $R(\matM_R, \beta)$ are computed by applying the learned models to the predicted parameters as explained in Eqs.~\ref{eq:model_geometry} and~\ref{eq:model_ref}. 
Thus, 
for each multi-frame image in a mini-batch, 
an expression deformation per sub-frame, 
and consistent identity and reflectance deformations across all sub-frames in the multi-frame image, are computed.
The computed reconstructions are then rendered using the differentiable renderer to produce synthetic images $S_i\in\mathbb{R}^{240X240X3}$.
We enforce orthogonality between the geometry and expression models to lead $\matM_{id} \matM_{exp} = \huge0$.
This is done by dynamically constructing $\matM_{id}$ in a forward pass by projecting itself onto the orthogonal complement of $\matM_{exp}$~\cite{Tewari19FML}.
Please see Fig.~\ref{fig:pipeline} for more details. 

\subsection{Dataset}
\label{Sec:Dataset}
\label{sec:method-dataset}
We use two datasets to train our approach: \emph{VoxCeleb}~\cite{Chung18b} and \emph{EmotioNet} \cite{BQ16}.
VoxCeleb is a multi-frame dataset consisting of over 140k videos covering 6000 different celebrities crawled from YouTube. 
The multi-frame nature of VoxCeleb allow us to train our Siamese network by feeding multiple frames (multi-frame images) for the same identity (see Fig.~\ref{fig:pipeline}). 
We sample 4 images per identity from the same video clip to avoid unwanted variations due to aging, accessories and so on. 
This gives us a variety of head pose, expressions and illumination per identity. 
All our images are cropped around the face, and we discard images containing less than 200 pixels. 
We resize the crops to 240x240 pixels. 

We also use EmotionNet~\cite{BQ16}. It is a large-scale still image dataset of in-the-wild faces, covering a wide variety of expressions, automatically annotated with Action Units (AU) intensities.
We use a subset of 7,000 images of neutral faces by selecting images with no active AU. 
We use these neutral faced images to enforce model disentanglement between the identity and expression geometry (Sec.~\ref{sec:ModelDisentanglement}). %

\subsection{Loss Functions}
We formulate several loss functions to train our network. 
We perform self-supervised training, without using any 3D supervision. 
Let $\mathbf{x}$ be the learnable variables in the network, which includes all trainable weights in the neural network, as well as the learnable face models $M_{id}$, $M_{exp}$ and $M_{R}$. 
All the estimated parameters $\vecp_i$ can be parametrized using these learnable variables.
Our loss function consists of:
\begin{align}
\label{eq:loss}
\nonumber
    \mathcal{L}(\mathbf{x}) = \mathcal{L}_{land}(\mathbf{x}) + \lambda_{seg}\cdot\mathcal{L}_{seg}(\mathbf{x}) +\\
    \nonumber
    \lambda_{pho}\cdot\mathcal{L}_{pho}(\mathbf{x}) +  \lambda_{per}\cdot\mathcal{L}_{per}(\mathbf{x}) +\\
    \lambda_{smo}\cdot\mathcal{L}_{smo}(\mathbf{x}) +  \lambda_{dis}\cdot\mathcal{L}_{dis}(\mathbf{x})
    \enspace{,}
\end{align}
The last two terms are regularizers and the first four are data terms. 
We used fixed $\lambda_\bullet$ values to weigh the losses.\\

\paragraph{Landmark Consistency}
\label{sec:landmark}
For each frame $F_i$, we automatically annotate $66$ sparse 2D keypoints~\cite{SaragihLC11a} $l\in\mathbb{R}^{66}$.
We compare these 2D landmarks with sparse vertices of the reconstruction which corresponds to these landmarks.
\begin{equation}
    \mathcal{L}_{land}(\mathbf{x}) = \sum_i^{N_f}\sum_{k=0}^{66} || l_k - p(v_k(\mathbf{x})) ||^2
    \enspace{.}
\end{equation}
Here, $v_k(\mathbf{x})\in\mathbb{R}^3$ indicates the position of the $k$th landmark vertex, and $p(v_k(\mathbf{x})$ is its 2D projection (Sec.~\ref{sec:image_formation}).
While most face landmarks can be manually annotated once on the template mesh, the face contour is not fixed and thus has to be calculated dynamically (see supplemental for details). 

\begin{figure}
\centering
    \includegraphics[width=0.8\linewidth]{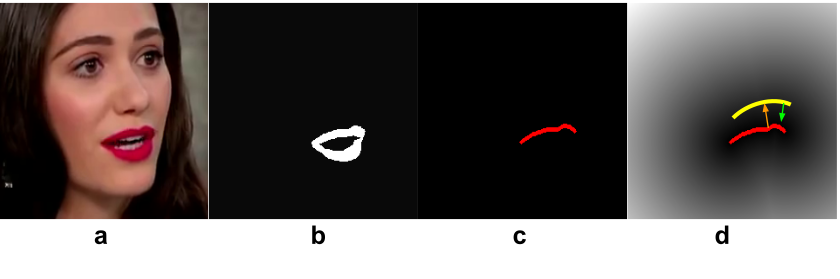}%
    \caption{For a given image [a], we obtain the segmentation masks [b], its boundary [c] and distance transform (DT) image [d] of [c]. We employ a segmentation loss which tries to move the vertices on the projected mesh contour (yellow) to a lower energy position in DT. In addition, each pixel in the boundary (red) attracts the nearest vertex on the mesh contour.}
    \vspace{-0.5cm}
	\label{fig:segAll}
\end{figure}

\paragraph{Segmentation Consistency} 
\label{sec:SegmentationConsistency}
The estimated keypoints are ambiguous in the inner lip regions, due to rolling lip contours. 
In addition, the accuracy of sparse keypoint prediction is inadequate to learn expressive expression models.
We use a dense contour loss for the lip region, guided by automatic segmentation mask prediction for the lips~\cite{CelebAMask-HQ}.
The lip segmentation contours are converted into distance transform images $\mathbf{D}_a^b$, where $a\in\{upper,lower\}$ and $b\in\{outer,inner\}$ corresponding to the outer and inner contours of both lips.
We also compute the contours of both lips projected by the predicted reconstruction $\mathbf{C}_a^b(x)$, where each element of $\mathbf{C}_a^b(x)$ stores a 2D pixel location. 
For a given distance transform image and the corresponding contour of the predicted mesh, the loss function minimizes the distance between the mesh contours and segmentation contours, see Fig.~\ref{fig:segAll}. 

\begin{figure}
\centering
    \includegraphics[width=0.8\linewidth]{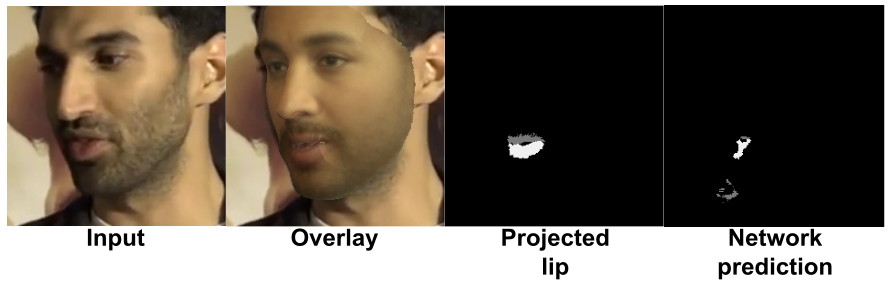}%
    \caption{At test time, our approach produces plausible upper (gray) and lower (white) lip segmentation even when the images are of bad quality, contain extreme poses or occlusions. For such cases ~\cite{CelebAMask-HQ} struggles to produce acceptable segmentation (column 4). }
    \vspace{-1cm}
	\label{fig:segRes}
\end{figure}

\begin{gather}
\label{loss:segmentation}
\nonumber
    \mathcal{L}_{seg}(\mathbf{x}) = \sum_{i=0}^{N_f}\sum_{\forall (a,b)} \, [ \sum_{\forall(x,y) \in \mathbf{C}_a^b(\mathbf{x})} \mathbf{D}_a^b(x,y)  + \\  \sum_{\{(x,y) \, | \, \mathbf{D}_a^b(x,y)=0\}} ||(x,y) - closest(\mathbf{C}_a^b(\mathbf{x}), (x,y))||^2 \, ]
    \enspace{.}
\end{gather}
Here, the first term minimizes the distance from every pixel in the mesh contour to the image contour. 
The second term is a symmetric term minimizing the distance between every pixel in the image contour to the closest mesh contour. $closest(\mathbf{C}_a^b(\mathbf{x}), (x,y))$ is a function which gives the position of the closest pixel in $\mathbf{C}_a^b$ to $(x,y)$.
We use our differentiable renderer to calculate $\mathbf{C}_l^{inner}(x)$ for the rolling inner contours. The outer contour $\mathbf{C}_l^{outer}(x)$ is computed as the projection of some manually annotated vertices on the template mesh. In practice, we ignore this  loss term at pixels where the distance between the image and mesh contours is greater than a threshold. This helps in training with noisy segmentation labels. 

\begin{figure*}
\centering
	\includegraphics[width=0.85\linewidth]{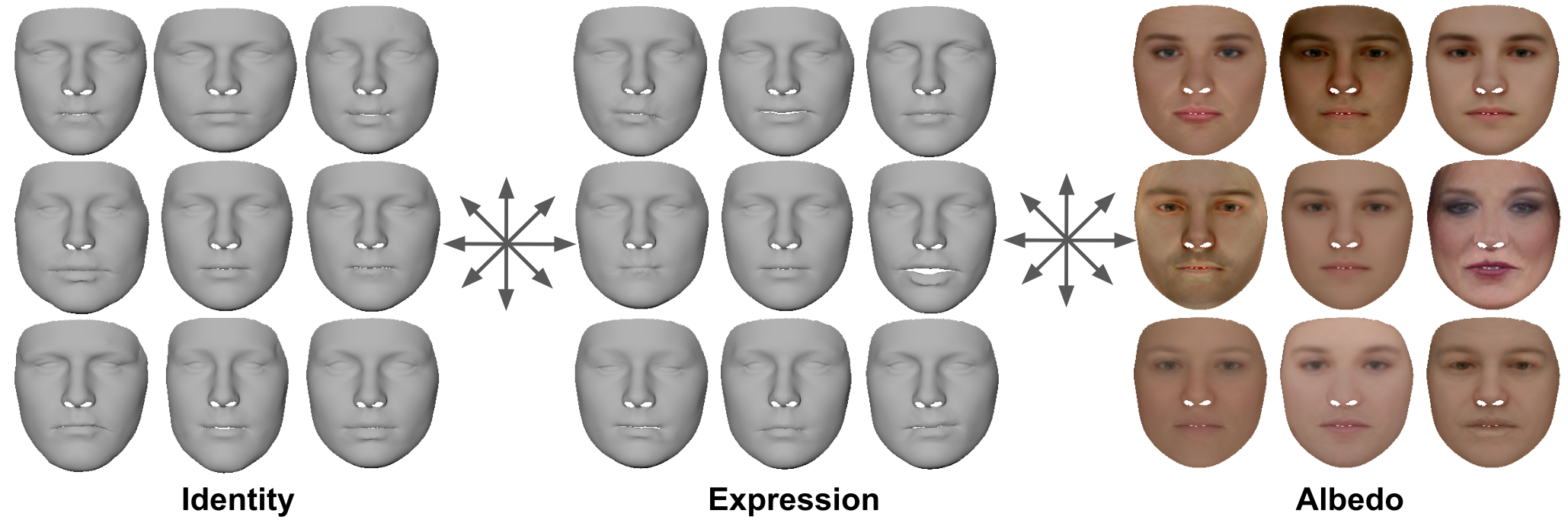}
	\caption{%
		Visualization of learned models. Faces in each direction of indicated arrows is obtained by linearly scaling individual component of respective models. Identity geometry captures variations in face shape (second column), lips (top left to bottom right) and jaw (top right to bottom left), while expressions capture variations due to mouth opening (second row), smile (second column) and eye movement (top right to bottom left). Albedo spans a variety of skin color (second column), eye color (top right to bottom left) and gender specific features e.g. facial hair, make-up (second row). 
	}
	\label{fig:ModelVisualization1}
\end{figure*}

\begin{figure*}
\centering
	\includegraphics[width=\linewidth]{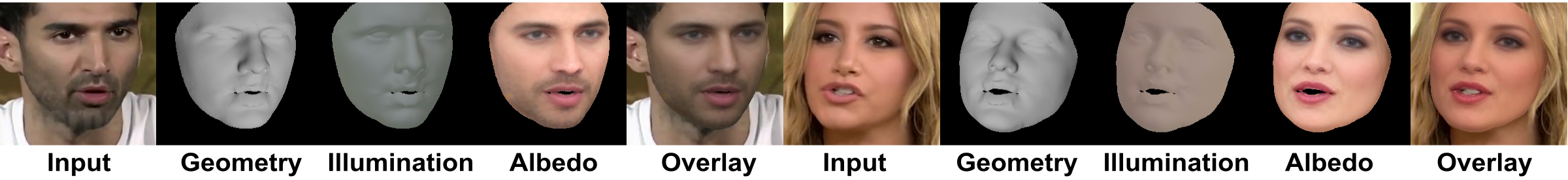}
	\caption{Our approach reconstructs all facial components with high fidelity and good disentanglement and results a photorealistic overlay. 
	}
	\label{fig:Ours}
	
\end{figure*}

\begin{figure*}
\centering
	\includegraphics[width=\linewidth]{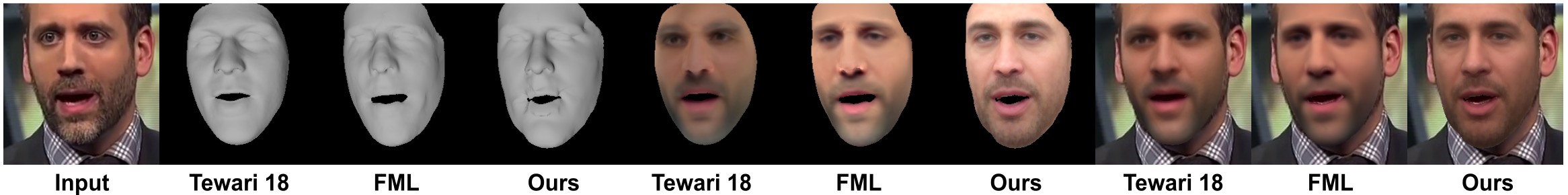}
	\caption{Our approach  produces better geometry, including detailed mouth shapes compared to Tewari~\etal~\cite{tewari17MoFA} and \emph{FML}~\cite{Tewari19FML}. Our albedo is also more detailed and better disentangled from the illumination.
	}
	\label{fig:SOTA5}
\end{figure*}

\begin{figure*}
\centering
	\includegraphics[width=\linewidth]{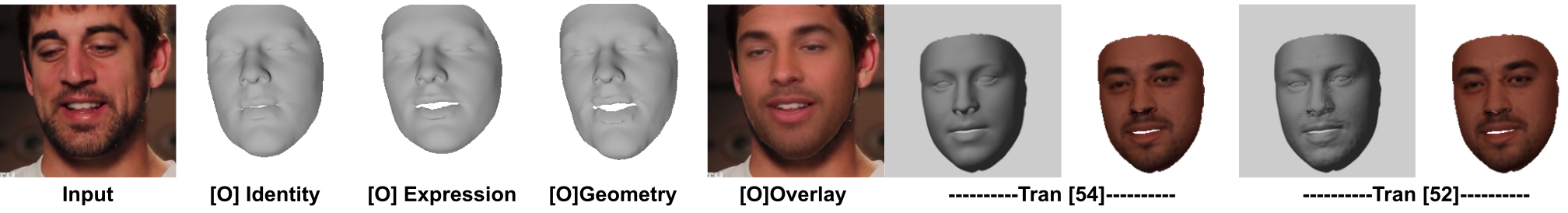}
	\caption{Both approaches of Tran~\etal~\cite{Tran18,Tran19} do not disentangle identity geometry from expressions. Our technique, however, estimates and disentagles all facial components. It also produces produces more accurate mouth shapes. 
	}
	\label{fig:SOTA2}
\end{figure*}

\paragraph{Photometric Consistency}
We evaluate the dense photometric consistency between the reconstructions and the input. 
For each pixel, we minimize the color difference between the input image $F$ and the rendered face $S$.
\begin{equation}
    \mathcal{L}_{pho}(x) = \sum_{i=0}^{N_f} || <M_i, (F_i - S_i(x))> ||^2
    \enspace{.}
\end{equation}
$M_i$ is a mask image with pixel value $1$ where the reconstructed mesh projects to. 

\paragraph{Perceptual Loss}
\label{loss:perceptual}
We additionally employ a dense perceptual loss to help our networks learn higher quality models, including high-frequency reflectance details. 
In particular, we use a VGG network pretrained on ImageNet~\cite{johnson2016perceptual} to get the intermediate features for both input frames and the output synthetic frames.
We then minimize the cosine distance between  these features. 
\begin{equation}
    \mathcal{L}_{per}(x) = \sum_{i=0}^{N_f} \sum_{l=0}^4 1 - \frac{ < f_l(S_i(x)) , f_l(F_i) >}{||f_l(S_i(x))|| \cdot ||f_l(F_i)||}
    \enspace{,}
\end{equation}
where $f_l(\cdot)$ denotes the output of the $l$th intermediate layer for input $x$ and $<\cdot,\cdot>$ denotes the inner product.

\paragraph{Geometry Smoothness}
To ensure smoothness of the final geometry, we use a smoothness loss at the graph level. 
Let $G_i \in \mathbb{R}^{N_g \times 3}$ with $N_g=521$  nodes denote the geometry reconstruction for frame $F_i$ at the graph level. 
We employ an $\ell2$ loss to constrain the difference between the deformation of adjacent nodes. 
\begin{equation}
    \mathcal{L}_{smo}(x) = \sum_{i=0}^{N_f} \sum_{g\in G_i} \sum_{n\in\mathcal{N}(g)}||g(x) - n(x)||^2
    \enspace{,}
\end{equation}
where $\mathcal{N}(g)$ is the neighbourhood of node $g$, $g(x)$ and $n(x)\in\mathbb{R}^3$.

\subsubsection{Model Disentanglement}
\label{sec:ModelDisentanglement}%
Our goal is to learn deformation models for facial geometry, expression and reflectance. 
Disentangling these deformations in the absence of an initial 3DMM is challenging.
We use a combination of network design choices and loss functions to enable simultaneous learning of these models. \\
\emph{Siamese Networks:} Our siamese network design ensures that the identity components of our reconstructions are consistent across all frames of the batch.
Such a network architecture allows us to disentangle illumination from reflectance in addition to helping with  disentanglement of  expressions from identity geometry.\\
\emph{Disentanglement Loss}
One example of a failure mode would be when $M_{id}$ collapses to a zero matrix.
Here, all geometric deformations including identity would be learned by the expression model without any penalty from any loss function.
To prevent such failure modes, we design a loss function to disentangle these components. 
As mentioned in~\ref{sec:method-dataset}, a subset of our dataset includes images which correspond to neutral expression. 
For these images, we employ a loss function which minimizes the geometry deformations due to expressions. 
\begin{equation}
    \mathcal{L}_{dis}(x) = \sum_{i=0}^{N_f}|| \delta_i(x) ||^2
    \enspace{.}
\end{equation}
Since we do not have videos for these images, we simply duplicate the same image as input to the siamese towers.
Finally, our training strategy further helps with disentanglement. Please refer to the supplemental for details.

\section{Results}
\label{sec:results}
\paragraph{Training Details} We implement our approach in \emph{Tensorflow} and train it over three stages: 1) pose pretraining 2) identity pretraining and 3) combined training. We empirically found this curriculum learning to help with stable training and disentanglement of the identity and expression models.
\emph{Pose Pretraining: }We first train only for the rigid head pose. All other parameters are kept fixed to their initial value. 
\emph{Identity pretraining: }Next, we train for the identity model. This step is only trained on the EmotionNet data with neutral expressions. 
We enforce the expression parameters to be zero, enforcing all deformations to be induced by the identity model. 
\emph{Combined Training: }Last, we train for the complete model with the loss functions as explained in \eqref{eq:loss}.
Similar to the first stage, we continue to impose the landmark loss term on the mean mesh throughout model learning. This helps in avoiding the geometric models learning the head pose.
Our training data now consists of mini-batches sampled from EmotionNet and VoxCeleb with 1:3 ratio. 
We train for $650$k iterations with a batch size of $1$. This results in a training time of $117$ hours on a TitanV. 
We use $80$ basis vectors for identity geometry and albedo, and $64$ for expression.

\begin{figure*}
\centering
	\includegraphics[width=\linewidth]{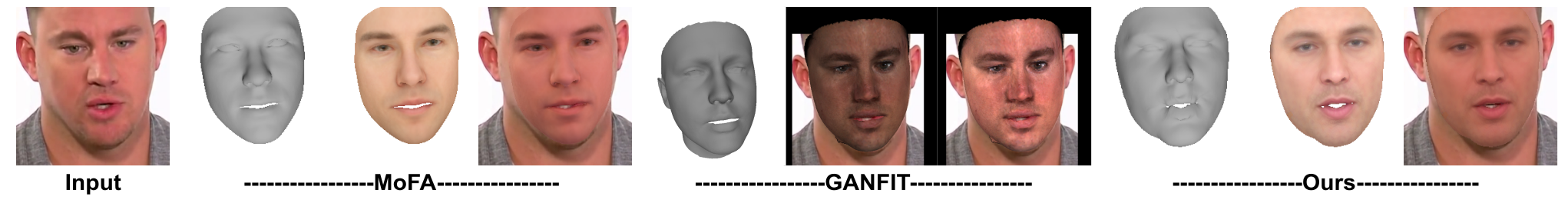}
	\vspace{-.4cm}
	\caption{
	MoFA~\cite{tewari17MoFA} and GANFIT~\cite{Gecer19CVPR} produce less accurate mouth shape than our technique. GANFIT~\cite{Gecer19CVPR} can produce artifacts in the albedo and final overlay, especially around the eyes. 
	}
	\label{fig:SOTA3}
\end{figure*}

\subsection{Qualitative Evaluation}
Fig.~\ref{fig:ModelVisualization1} visualizes the different modes of learned model. Our method disentangles the various facial components of identity geometry, expressions and albedo.  
The identity model correctly captures a variety of face shapes, mouth and eye structure. The expression model captures a variety of movements produced by the mouth and eyes, while the albedo captures different skin color, and gender specific features such as facial hair and make-up.
Fig.~\ref{fig:Ours} shows all components of our reconstruction for several images.
Our approach can handle different ethnicities, genders and scene conditions, and produces high-quality reconstruction, both in geometry and reflectance.

\begin{figure*}
\centering
    \includegraphics[width=\linewidth]{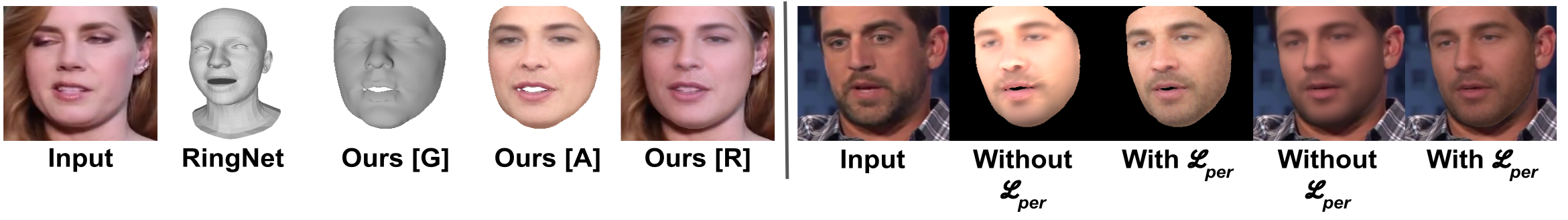}%
    \caption{Left: Our technique better captures mouth shape and eye geometry than \emph{RingNet}~\cite{RingNet19}. It also produces a photorelistic overlay. Right: Albedo and overlay is noticeably improved with perceptual loss $\mathcal{L}_{per}$ (see 3rd and 5th column).}
	\label{fig:SOTA4_PerceptualLoss}
\end{figure*}

Fig.~\ref{fig:segRes} shows that our method can produce better lip segmentation than the approach used for generating the training data~\cite{CelebAMask-HQ} in some cases. This is due to our segmentation loss function, $\mathcal{L}_{seg}$, where we selectively ignore unreliable segmentation estimates. Hence our final model is learned from only accurate segmentations in the training-set. 
Fig.~\ref{fig:SOTA4_PerceptualLoss} shows that including perceptual loss in our training (Sec.~\ref{loss:perceptual}) clearly improves the photorealism of the albedo and the final overlay. 
\emph{Comparisons: }
Fig.~\ref{fig:SOTA2} - \ref{fig:SOTA4_PerceptualLoss} compare our approach to several state-of-the-art face reconstruction techniques.
Tran~\etal~\cite{Tran18,Tran19} learns a combined geometry for identity and expressions, while we learn a separate model for each (Fig.~\ref{fig:SOTA2}). 
\emph{MoFA}~\cite{tewari17MoFA} and \emph{GANFIT}~\cite{Gecer19CVPR} geometry are limited by a pretrained 3DMM model and hence lead to less detailed shapes than ours (Fig.~\ref{fig:SOTA3}). While \emph{GANFIT} produces detailed textures, it can contain artifacts. Like other 3DMM based approaches, \emph{RingNet}~\cite{RingNet19}, which estimates the parameters of a pre-trained face model~\cite{FLAME017}, struggles with out-of-space variations especially in the mouth region (Fig.~\ref{fig:SOTA4_PerceptualLoss}). Tewari~\etal~\cite{tewari2018self} refines a pretrained 3DMM model on an image dataset. We can disentangle the reflectance and illumination better (Fig.~\ref{fig:SOTA5}).
\emph{FML}~\cite{Tewari19FML} is constrained by a pretrained expression model and thus produces less convincing shape reconstructions than us (Fig.~\ref{fig:SOTA5}). 
In addition, our reflectance estimates are more detailed compared to~\cite{Tewari19FML}.
Thus, even though we start from just a template mesh without any deformation priors, we can produce high-quality results, better than the state-of-the-art. 

\subsection{Quantitative Evaluation}
\emph{Geometric Error: } To evaluate the geometric accuracy of our 3D reconstructions, we  compute the per-vertex root mean square error between the ground-truth geometry and the geometry estimated using different techniques. 
We evaluate this metric on the BU3DFE dataset~\cite{Yin06} where the  ground-truth geometry is obtained using 3D scans.
Tab.~\ref{table:BU3DError} reports the results over $324$ images. 
Our approach outperforms the approaches of MoFA~\cite{tewari17MoFA},  Tewari~\etal~\cite{tewari2018self} and \emph{FML}~\cite{Tewari19FML}.
Note that none of the approaches in Tab.~\ref{table:BU3DError} learn a complete face model from images and videos.
\\
\emph{Segmentation Error: } To specifically evaluate the quality of lip reconstructions, we use Intersection over Union (IoU) between our reconstructions and the input images over the lip regions. 
Since our approach learns an expression model from in-the-wild data, it can generalize better to different lip shapes and  significantly outperform \emph{FML}~\cite{Tewari19FML} (see Tab.~\ref{table:SegmentationErrorAblation2}). 
Furthermore, Tab.~\ref{table:SegmentationErrorAblation2} shows that removing the segmentation consistency term (Eq.~\ref{loss:segmentation}) leads to lower quality results.\\
\emph{Disentanglement Error: } 
One of our main objectives is to obtain a disentangled representation for faces. 
To evaluate the disentanglement between the reconstructed expression and identity geometry, we design a metric which measures the average of expression deformation for images with neutral faces. 
We tested our approach on $1864$ neutral faces mined using the same strategy described in Sec.~\ref{Sec:Dataset}. Tab.~\ref{table:IdentityAblation2} reports the average $\ell_2$ value of the expression deformations, as estimated using different approaches. The lower the expression strength, the better the disengagement. Our approach achieves significantly better expression and identity disentanglement over FML~\cite{Tewari19FML} and MoFA~\cite{tewari17MoFA}.\\ 
\emph{Verification Metric:}
To further evaluate disentanglement, we use the LFW dataset \cite{LFWTech}, which includes face image pairs of same as well as different identities. 
We render the identity component of the reconstructions with the predicted pose and lighting parameters. 
Face embeddings are computed as the average pooled version of \texttt{conv5\_3} output of  VGG-Face~\cite{Parkhi15}. 
We first compute the histogram distribution of cosine similarities between renderings of image pairs with the same identity in embedding space. 
Similarly, distribution of cosine similarities between rendering pairs of different identities is computed. 
Then, we compute the verification metric as the Earth Movers Distance (EMD) between these two distributions.
Our method achieves an EMD of $0.15$, compared to $0.09$ of FML. 
A larger distance implies that the network can better represent the difference between different identities due to better disentanglement.

\begin{table}[]
\centering
\begin{tabular}{lllllll}
\cline{1-6}
\multicolumn{1}{|l|}{}     & \multicolumn{1}{l|}{Ours} & \multicolumn{1}{l|}{MoFA} & \multicolumn{1}{l|}{FML}  & \multicolumn{1}{l|}{Fine~\cite{tewari2018self}}  & \multicolumn{1}{l|}{Coarse~\cite{tewari2018self}}  & \\ \cline{1-6} %
\multicolumn{1}{|l|}{Mean} & \multicolumn{1}{l|}{\textbf{1.75}}     & \multicolumn{1}{l|}{3.22}       & \multicolumn{1}{l|}{1.78}       & \multicolumn{1}{l|}{1.83}       &\multicolumn{1}{l|}{1.81}            & \\ \cline{1-6} %
\multicolumn{1}{|l|}{SD}   & \multicolumn{1}{l|}{\textbf{0.44}}     & \multicolumn{1}{l|}{0.77}       &
\multicolumn{1}{l|}{0.45}          &
\multicolumn{1}{l|}{0.39}      &   \multicolumn{1}{l|}{0.47}      &     \\ \cline{1-6} %
                           &                           &                             &                                                & & &   
\end{tabular}
\caption{Geometric reconstruction error (in mm) on the BU-3DFE dataset~\cite{Yin06}. 
Our technique outperforms \emph{MoFA}~\cite{tewari17MoFA}, coarse and fine models of Tewari~\etal~\cite{tewari2018self} and  \emph{FML}~\etal~\cite{Tewari19FML}.
}
\label{table:BU3DError}
\end{table}

\begin{table}
\centering
\begin{tabular}{lllll}
\cline{1-4}
\multicolumn{1}{|l|}{}    & \multicolumn{1}{l|}{W/o $\mathcal{L}_{seg}$} & \multicolumn{1}{l|}{With $\mathcal{L}_{seg}$} & \multicolumn{1}{l|}{FML} &  \\ \cline{1-4}
\multicolumn{1}{|l|}{UL IoU} & \multicolumn{1}{l|}{0.49}                          & \multicolumn{1}{l|}{\textbf{0.52}}  & \multicolumn{1}{l|}{0.51}                      &  \\ \cline{1-4}
\multicolumn{1}{|l|}{LL IoU} & \multicolumn{1}{l|}{0.53}                          & \multicolumn{1}{l|}{\textbf{0.61}}   & \multicolumn{1}{l|}{0.56}                     &  \\ \cline{1-4}
&
&
&
\end{tabular}
\caption{Intersection over Union (IoU) between ground-truth lip mask and the segmentation produced by different techniques. Our segmentation consistency term produces better IoU and leads to noticeably better performance than \emph{FML}~\cite{Tewari19FML}}
\label{table:SegmentationErrorAblation2}
\end{table}

\begin{table}
\centering
\begin{tabular}{llllll}
\cline{1-5}
\multicolumn{1}{|l|}{}    & \multicolumn{1}{l|}{W/o $\mathcal{L}_{dis}$ } & \multicolumn{1}{l|}{With $\mathcal{L}_{dis}$} & \multicolumn{1}{l|}{FML} & \multicolumn{1}{l|}{MoFA} &  \\ \cline{1-5}
\multicolumn{1}{|l|}{AE} & \multicolumn{1}{l|}{2.5075}                          & \multicolumn{1}{l|}{\textbf{0.0116}}      & \multicolumn{1}{l|}{2.0329}    & \multicolumn{1}{l|}{0.4056}              &  \\ \cline{1-5}
                          &                                                &                                             & &
\end{tabular}
\caption{Our identity disentanglement term results in lesser leakage of identity geometry into expression component in neutral faces. It  performs better than \emph{FML}~\cite{Tewari19FML} and \emph{MoFA}~\cite{tewari17MoFA}}
\label{table:IdentityAblation2}
\end{table}

\section{Conclusion}
\label{sec:discussion}
We presented the first approach for learning a full face model, including learned identity, reflectance and expression models from in-the-wild images and videos.
Our method also learns to reconstruct faces on the basis of the learned model from monocular images. 
We introduced new training losses to enforce disentanglement between identity geometry and expressions, and to better capture detailed mouth shapes. 
Our approach outperforms existing methods, both in terms of the quality of image-based reconstruction, as well as disentanglement between the different model components. 
We hope that our work will inspire further research on building 3D models from 2D data.

\paragraph{Acknowledgments.} 
This work was supported by the ERC Consolidator Grant 4DReply (770784). 

{\small
\bibliographystyle{ieee}
\bibliography{refs}
}

\end{document}